\documentclass[journal]{IEEEtran}
\usepackage{graphicx}
\usepackage{amsmath, bm}
\usepackage{wrapfig}
\usepackage{amsmath}
\usepackage{multirow, array}
\usepackage{algorithmic}
\usepackage[noadjust]{cite}
\usepackage{color}

\begin{document}
%
\title{Deep Neural Networks for Surface Segmentation Meet Conditional Random Fields}
%
%
%

\author{Leixin~Zhou,
        Zisha~Zhong,
	Abhay~Shah,
 Bensheng Qiu, John Buatti, 
        and~Xiaodong~Wu,~\IEEEmembership{Senior~Member,~IEEE}
\thanks{
L. Zhou and X. Wu are with the Department of Electrical and Computer Engineering,
The University of Iowa, Iowa City, IA 52242, USA (e-mail: leixin-zhou,xiaodong-wu@uiowa.edu). 

Z. Zhong and A. Shah were with the Department of Electrical and Computer Engineering,
The University of Iowa, Iowa City, IA 52242, USA (e-mail: zisha-zhong, abhay-shah-1@uiowa.edu).

B. Qiu is with School of Information Science and Technology, The University of Science and Technology of China, Hefei 230027, China (e-mail: bqiu@ustc.edu.cn).

J. Buatti is with the Department of Radiation Oncology, The University of Iowa, Iowa City, IA 52242, USA (e-mail: john-buatti@uiowa.edu).
}}

\maketitle

\begin{abstract}
Automated surface segmentation is important and challenging in many medical image analysis applications. Recent deep learning based methods have been developed for various object segmentation  tasks. Most of them are a classification based approach (e.g., U-net), which predicts the probability of being target object or background for each voxel. One problem of those methods is lacking of topology guarantee for segmented objects, and usually post processing is needed  to infer the boundary surface of the object. In this paper, a  novel model based on $3$-D convolutional neural networks (CNNs) and Conditional Random Fields (CRFs) 
is proposed to tackle the surface segmentation problem with end-to-end training. To the best of our knowledge, this is the first study to apply a $3$-D neural network with a CRFs model for direct surface segmentation. Experiments carried out on NCI-ISBI 2013 MR prostate dataset and Medical Segmentation Decathlon Spleen dataset demonstrated promising segmentation results.

\end{abstract}


\IEEEpeerreviewmaketitle

\section{Introduction}

\IEEEPARstart{A}{utomated}
image segmentation plays an essential role in quantitative image analysis. Semantic segmentation methods based on convolutional neural networks (CNNs) have grown in popularity in both the computer vision and medical imaging research communities. Fully convolutional networks (FCNs) \cite{long2015fully} are applied to natural images, while U-net \cite{ronneberger2015u} and its $3$-D version V-net \cite{milletari2016v} are used for medical image segmentation. 
\par
As pixels or voxels usually exhibit strong correlation in both natural and medical images, jointly modeling the global and local label distribution between them is desirable. To capture the contextual information, conditional random fields (CRFs)~\cite{jordan1999introduction} are commonly utilized for semantic segmentation. The model consists of a unary and a pairwise potential term. The unary potential specifies the per-pixel or voxel confidence of assigning a label, while the pairwise potential regularizes the label smoothness between neighboring voxels.  In computer vision, the CRFs model was integrated with CNNs for an end-to-end training to take advantage of both the modeling power of CRFs and the representation-learning ability of CNNs~\cite{zheng2015conditional}.
\par
Most deep learning based semantic segmentation methods are {\em classification} or {\em region} based \cite{long2015fully,ronneberger2015u,milletari2016v,zheng2015conditional}, in which each pixel is labeled as either target object or background. On the other hand, one can also formulate semantic segmentation with a {\em surface-based} model, in which the boundary surface of the target object is computed directly. These two types of approaches are equivalent as the boundary surface can be computed from the labeled target volume, and vice versa. The Graph-Search (GS) \cite{wu2002optimal,li2006optimal} method is one of the prominent {\em surface-based} methods where it has been widely used in the medical imaging field, \cite{garvin2009automated, yin2010logismos, oguz2014logismos,garvin2008intraretinal, song2013optimal}. This method is capable of simultaneously detecting multiple interacting surfaces of global optimality with respect to the energy function designed for the target surfaces with geometric constraints, which define the surface smoothness and interrelations. The method solves the surface segmentation problem by transforming it to compute a minimum \textit{s-t} cut in a derived arc-weighted directed graph, which can be solved optimally with a low-order polynomial time complexity.
\par
Inspired by the GS method, Shah \textit{et al.}  \cite{shah2017simultaneous,shah2018multiple} first modeled the terrain-like surfaces segmentation as direct surface identification using a regression network based on CNNs. The network only models the unary potentials. As the prediction was directly on surface positions, a surface monotonicity constraint was realized in a straightforward way. The network used was a very light-weighted $2$-D CNN and no post processing was required. Surprisingly the results were very promising. It would be of high interest to extend Shah \textit{et al.}'s method to segment $3$-D  general non-terrain like surface. To achieve this goal, two major obstacles need to be overcome: 1) how to generate patches with a regular neighborhood in $3$-D, such that the traditional CNNs can be applied? 2) how to train a sufficient number of paramaters? It is generally hard to train a $3$-D network, especially when it contains giant fully connected (FC) layers, the size of which is closely related to inference/patch size. There is a tradeoff between the amount of contextual information within a patch and the number of parameters in a network architecture, i.e. a bigger patch size comes along with more contextual information, but more parameters need to be trained. 
\par
{\em Contributions}:
To overcome those technical barriers, we propose building a framework of {\em surface-based} CNN+CRFs for surface segmentation in medical images. The framework strives to properly model the CRF unary term and the CRF pairwise term within the deep neural network with customized compatibility matrices for surface segmentation. In addition, we propose a novel shape-aware patch generation method, which is based on harmonic mapping, to make efficient training of surface segmentation possible. 
\section{Method}
\begin{figure}

    \centering
    \includegraphics[width=0.9\linewidth]{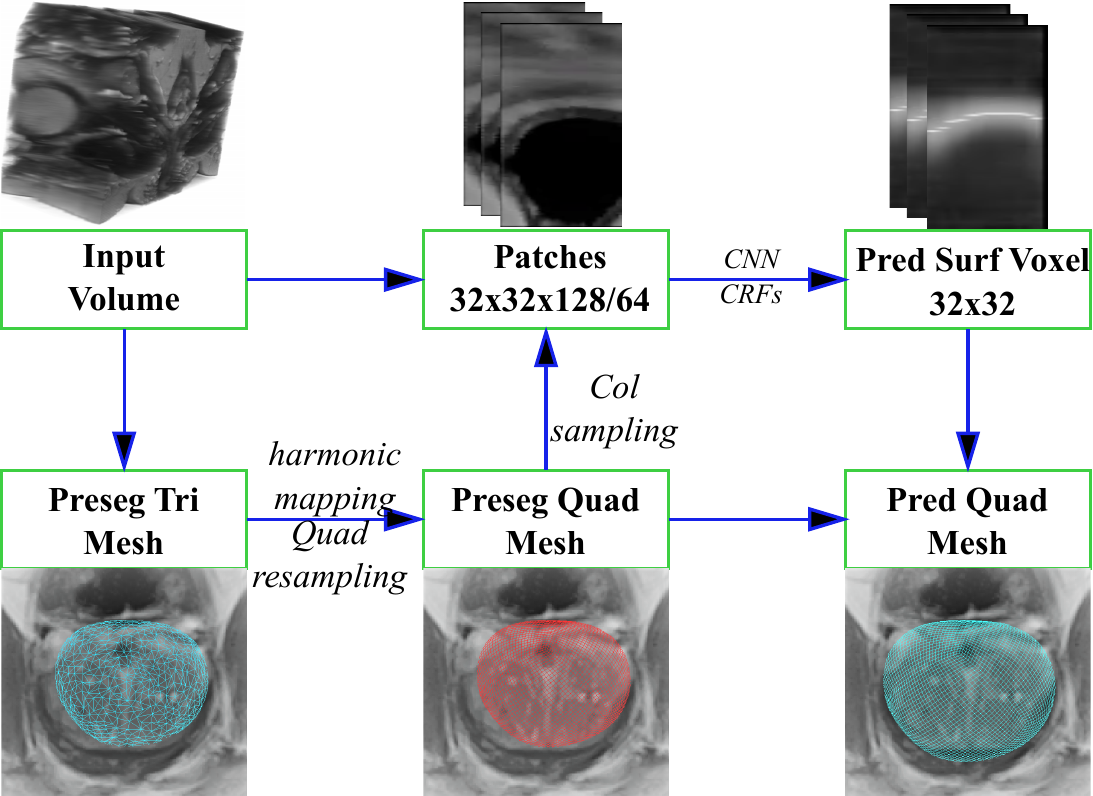}
    \caption{Overall flowchart of the proposed method. With some pre-segmentation method, a coarse surface segmentation or pre-segmentation ({\em preseg}) can be generated and represented as a triangular ({\em tri}) mesh. Then the proposed remeshing method would convert the {\em preseg} {\em tri} mesh into a {\em preseg} quadrilateral ({\em quad}) mesh. Based on the {\em preseg} {\em quad} mesh, for each vertex in which, sampling within input volume in its normal direction with fixed resolution and length would generate a column. Combining the columns for all vertices on the {\em preseg} {\em quad} mesh produces the $3$-D volumes/patches, which are the inputs for the proposed network. Then the network predicts surface voxels, which finally gives the prediction ({\em pred}) {\em quad} mesh surface.}
    \label{fig:flow}
\end{figure}

The pipeline of the proposed method starts with a pre-segmentation ({\em preseg}), which serves as the rough surface position and topology that the final segmentation should comply with. The triangular ({\em tri}) mesh of the {\em preseg} surface is then converted to a quadrilateral ({\em quad}) mesh, which is friendly to convolution operations. Based on the {\em quad} mesh, image patches, which contain terrain-like boundary surfaces of the partial target object, can be generated and are fed into the proposed neural network to predict the voxels on the desired surface. The flowchart of the proposed method is illustrated in Fig.~\ref{fig:flow}. 
\par 
In the following sections, we will first define the {\em surface-based} segmentation problem rigorously, followed by a brief review of the CRFs model. The modeling of the unary and pairwise terms will then be discussed. Finally, a novel shape-aware patch generation method and the network architecture will be presented. 

\subsection{Surface-Based Segmentation}
\begin{figure}[h]
    \centering
    \includegraphics[width=0.75\linewidth]{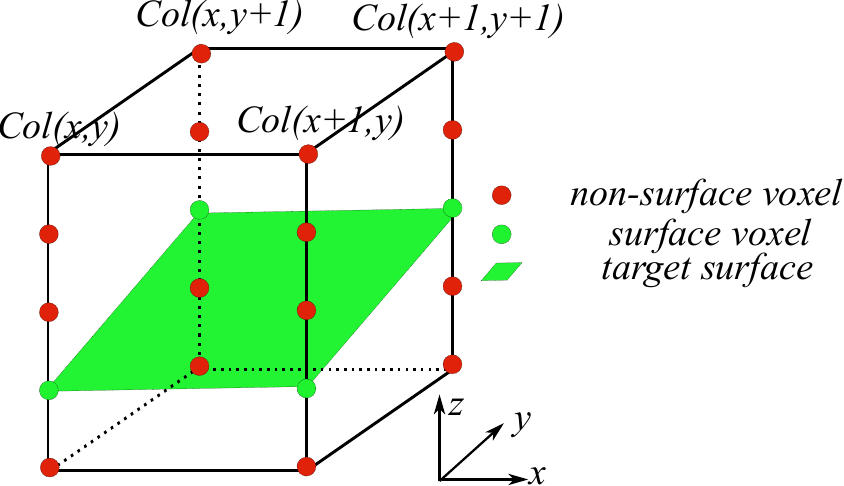}
    \caption{Definition of the surface segmentation for volumetric images.}
    \label{fig:surf_def}
\end{figure}
\par
A $3$-D image can be viewed as a $3$-D tensor $\mathcal{I}(x, y, z)$. A terrain-like {\em surface} in $\mathcal{I}$ is oriented and shown in Fig.~\ref{fig:surf_def}. Let $X$, $Y$ and $Z$ denote the image sizes in the $x$, $y$ and $z$ dimensions, respectively. The surface is defined by a function $\mathcal{N}: (x, y)$ $\rightarrow$ $\mathcal{N}_{x, y}$, where $\mathbf{x} = \{0, \ldots, X-1\}$, $\mathbf{y} = \{0, \ldots, Y-1\}$, and $\mathbf{z} = \{0, \ldots, Z-1\}$. Thus any surface in $\mathcal{I}$ intersects with exactly one voxel of each {\em column} ({\em Col}) in parallel with the $z$ direction, and it consists of exactly $X \times Y$ voxels. The optimal surface segmentation problem is formulated as solving the optimization problem:
\begin{equation}
\mathcal{N} = \text{arg}\min_{\mathcal{N}} E_u(\mathcal{N} | \mathcal{I}) + w_p E_p(\mathcal{N} | \mathcal{I}).
\label{surf-model}
\end{equation}
 The unary term $E_u$ in the energy function ensures the target surface follows the prominent image features. The pairwise energy term $E_p$ penalizes the discontinuity of the surface positions among adjacent columns to enforce surface smoothness. The coefficient $w_p$  balances the contribution of the two terms.
\subsection{CRFs Model}
 CRFs is defined on observations $\mathcal{X}$ and random variables $\mathcal{Y}$, as follows. Let $\mathcal{G} = (\mathcal{V}, \mathcal{E})$ be a graph. $\mathcal{Y} = \{\mathcal{Y}_v | v \in \mathcal{V}\}$ is a set of random variables defining on the vertices of $\mathcal{G}$, and $\mathcal{X}$ is observations over $\mathcal{V}$. Then $(\mathcal{X}, \mathcal{Y})$ is a conditional random field when the random variables $\mathcal{Y}_v$ conditioned on $\mathcal{X}$ while obeying  the Markov property.
The pair $(\mathcal{X}, \mathcal{Y})$ can be  characterized by a Gibbs distribution of the form 
 \begin{equation}
 P(\mathcal{Y} = \bm{y} | \mathcal{X}) = \frac{1}{\mathcal{Z}(\mathcal{X})}\text{exp}(-E(\bm{y}|\mathcal{X})).
 \end{equation}
  Here $E(\bm{y}|\mathcal{X})$ is  the energy of the configuration $\bm{y}$ and $\mathcal{Z}(\mathcal{X})$ is the partition function. For convenience, the conditioning on $\mathcal{X}$ is dropped. In a fully connected pairwise CRFs model, the energy of a label assignment $\bm{y}$ is given by:
\begin{align}
\label{eq:crf_def}
E(\bm{y}) &= \sum_{v} \psi_{u}(\bm{y}_v) + w_p\sum_{(v,v')\in \mathcal{E}} \psi_{p}(\bm{y}_v, \bm{y}_{v'}),
\end{align}
where the unary energy component $\psi_{u}(\bm{y}_v)$ measures the energy of  $\mathcal{Y}_v$ taking the label $\bm{y}_v$, and pairwise energy component $\psi_{p}(\bm{y}_v, \bm{y}_{v'})$ measures the energy of assigning $\bm{y}_v$, $\bm{y}_{v'}$ to  $\mathcal{Y}_v$ and $\mathcal{Y}_{v'}$ simultaneously. $w_p$ is the coefficient to balance the two terms.
\subsection{Modeling the Surface Segmentation as CRFs}
It is  natural to model the {\em surface-based} segmentation with a CRFs model. In the {\em surface-based} segmentation scenario, $\mathcal{N}$ are the random variables and $\mathcal{I}$ is the observation.
The unary potential $\psi_{u}(n_{x,y})$ corresponds to the energy of assigning the surface position  $n_{x,y}$ to $\mathcal{N}_{x,y}$ of the column $Col_{x,y}$.  Note that $n_{x,y} \in \mathbf{z} = \{0,1, \ldots, Z-1\}$ forms the label set in the CRFs model. The pairwise potential $\psi_{p}(n_{x,y}, n_{x',y'})$ represents the energy to simultaneously assign surface positions  $n_{x,y}$ and $n_{x',y'}$, respectively, to $\mathcal{N}_{x, y}$ and $\mathcal{N}_{x', y'}$ of two adjacent $Col_{x,y}$ and $Col_{x', y'}$. 
\par 
We propose using CNN, as in semantic segmentation~\cite{krahenbuhl2011efficient}, to compute the unary potentials $\psi_u(n_{x,y})$ for each column $Col_{x,y}$, thus obtaining the unary energy term $E_u(\mathcal{N} | \mathcal{I}) = \sum_{(x, y) \in \mathbf{x} \times \mathbf{y}}\psi_u(n_{x,y})$ for the target surface $\mathcal{N}$. 
\par 
The pairwise energy term $E_p(\mathcal{N} | \mathcal{I})$ is the total sum of the pairwise potentials $\psi_p(n_{x, y}, n_{x', y'})$ over all adjacent columns $Col_{x,y}$ and $Col_{x', y'}$ to encourage assigning similar surface positions to adjacent columns with similar properties. The pairwise potential can be modeled as weighted Gaussians~\cite{krahenbuhl2011efficient}, as follows. 
\begin{align}
 \psi_{p}(n_{x,y}, n_{x',y'}) &= \mu(n_{x,y}, n_{x',y'}) k(\mathrm{f}_{x,y}, \mathrm{f}_{x',y'}) \nonumber \\
= \mu(n_{x,y}, n_{x',y'}) & \sum_{m=1}^{M} g_m(\mathrm{f}_{x,y}, \mathrm{f}_{x',y'}),
\end{align}
where each $g_{m}$ for $m=1, ..., M$, is a Gaussian kernel applied on the feature vectors $\mathrm{f}_{x,y}$ and $\mathrm{f}_{x',y'}$. The feature vectors  $\mathrm{f}_{x, y}$ of $Col_{x,y}$ are derived from image features such as spatial location, and visual features like pixel/voxel intensities. The  label compatibility function $\mu$ captures the compatibility between different pairs of surface positions for adjacent columns.
\par
In~\cite{krahenbuhl2011efficient}, the term $k(\mathrm{f}_{x,y}, \mathrm{f}_{x',y'})$ is defined as:
\begin{equation}
\begin{split}
k(\mathrm{f}_{x,y}, \mathrm{f}_{x',y'}) = \text{exp}\left(-\frac{||(x'-x,y'-y)||^2}{2\theta_{1}^{2}}\right. \\
-\left.\frac{||Col_{x,y}-Col_{x',y'}||^2}{2\theta_{2}^{2}}\right) \\
+  w_1\text{exp}\left(-\frac{||(x'-x,y'-y)||^2}{2\theta_{3}^{2}}\right),
\end{split}
\label{eq:kernel}
\end{equation}
where the first and second terms are an  \textit{appearance kernel} and a \textit{smoothness kernel}, respectively.  $\theta_1$, $\theta_2$, and $\theta_3$ control the shapes of the corresponding Gaussian kernels.
\par
\begin{figure}[h]

    \centering
    \includegraphics[width=1.0\linewidth]{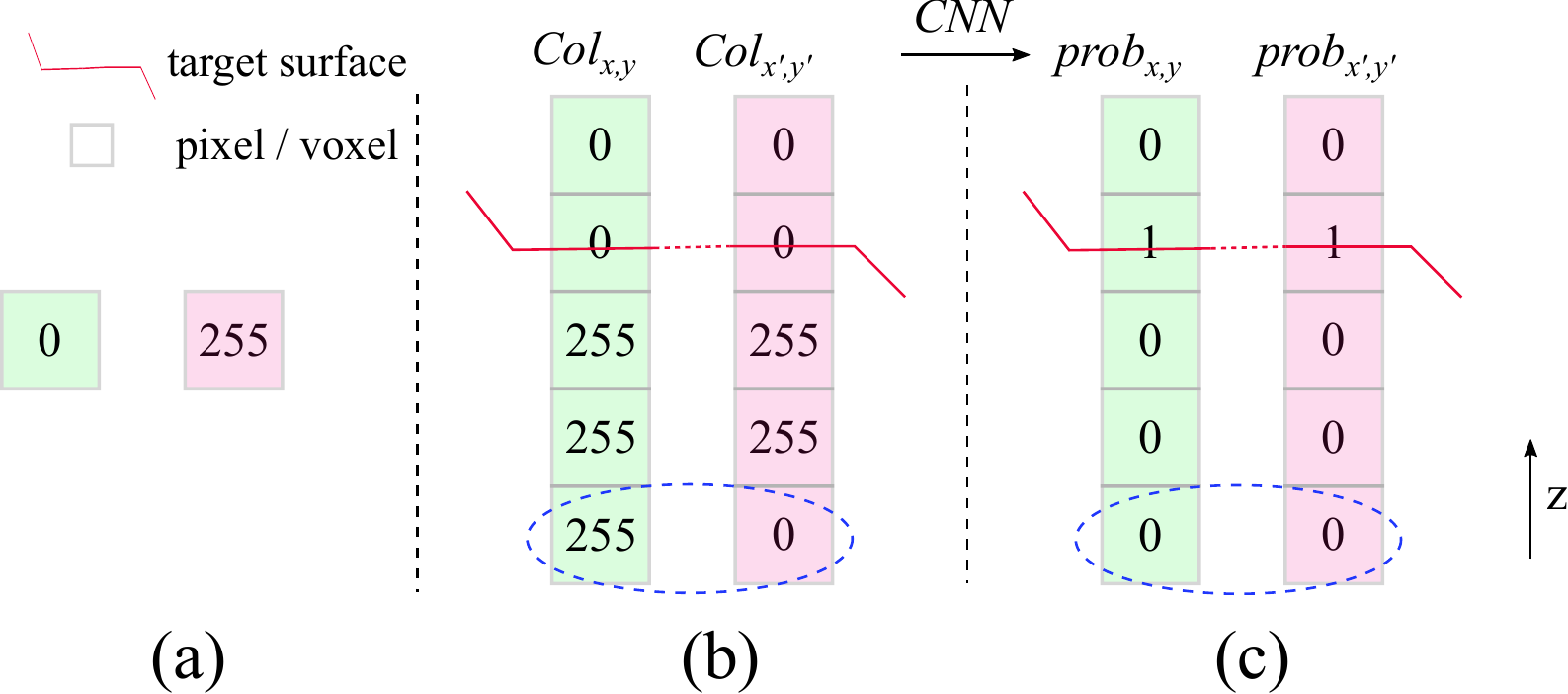}

    \caption{Visual Features in different settings. For convenience, gray intensity value is shown. (a) In $2$-D {\em region-based} segmentation, $Col_{x,y}$ is just a scalar intensity value. (b) In $3$-D {\em surface-based} segmentation, $Col_{x,y}$ is a column of voxels in the $z$ dimension. It can be noticed that within columns, intensity differences of voxel pairs that are not around the target surface actually does not relate much to the labellings of the column pairs. One voxel pair is outlined by the blue dash oval. (c) From our observation, the probability map or logits of each column is more proper to use as visual features, as the CNN may remove the interference of unrelated voxel pairs by having a global view.}

    \label{fig:visual_feature}
\end{figure}

 
Unfortunately, $k(\mathrm{f}_{x, y}, \mathrm{f}_{x', y'})$ defined in Eqn.~(\ref{eq:kernel}) may not be appropriate for our surface segmentation setting. In $2$-D {\em region-based} segmentation in computer vision~\cite{krahenbuhl2011efficient}, $Col_{x,y}$ is a single intensity value in a gray image or is a RGB vector in a color image. Thus, $||Col_{x,.y} -Col_{x',y'}||^2$ measures the appearance difference of two adjacent pixels $(x, y)$ and $(x', y')$, which may indicate their possible label difference. However, in our {\em surface-based} segmentation setting, $Col_{x,y}$ represents one column of voxels in the input $3$-D image $\mathcal{I}$. It thus contains mixed information of other structures. The voxels away from the one on the target surface may not play a significant role for determining the surface position. While computing $||Col_{x,.y} -Col_{x',y'}||^2$, those voxels may contribute large variance, e.g., the two voxels $Col_{x,y}[0]$ and $Col_{x',y'}[0]$ outlined by the blush dash ovals in Fig.~\ref{fig:visual_feature} (b).
To remedy this problem, we propose using  {\em probability-map} or {\em logits} output by the CNN as the visual features (Fig.~\ref{fig:visual_feature} (c)). 
The new kernel term is of the form
\begin{equation}
\begin{split}
k(\mathrm{f}_{x,y}, \mathrm{f}_{x',y'}) = \text{exp}\left(-\frac{||(x'-x,y'-y)||^2}{2\theta_{1}^{2}}\right. \\
-\left.\frac{||prob_{x,y}-prob_{x',y'}||^2}{2\theta_{2}^{2}}\right) \\
+  w_1\text{exp}\left(-\frac{||(x'-x,y'-y)||^2}{2\theta_{3}^{2}}\right).
\end{split}
\label{eqn-kernel}
\end{equation}
\par 
In $2$-D {\em region-based} semantics segmentation, the nature of the difference between classes does not have a fixed or absolute meaning. Suppose classes {\em cat}, {\em car} and {\em building} have labels of 0, 1 and 2, respectively. There is no way to append meaning to the label difference as there is no denotative system for describing how each label is structurally different. A {\em cat} is merely a label with no inherent defining qualities that separate its meaning from say {\em building} or {\em car}. In contrast, in our $3$-D {\em surface-based} segmentation, the surface position difference between adjacent columns has an explicit meaning of {\em surface smoothness}: the smaller the position difference, the smoother the surface.  
\par
A straightforward way to learn the compatibility matrix would be to learn a $Z \times Z$ matrix. In our scenario, that is ill-posed, since some position pairs may not exist or are sparsely presented in the training data. To tackle that issue, we propose to parameterize the compatibility matrix ($Z \times Z$) with a parameter function $\mathcal{C}$, as follows. 
\begin{equation}
\begin{split}
\mu(n_{x,y}, n_{x',y'}) = \mathcal{C}\left(|n_{x,y} - n_{x',y'}|\right) \\
=-\text{exp}\left(-\frac{(n_{x,y} - n_{x',y'})^2}{\theta_{\text{comp}}^2}\right).
\end{split}
\end{equation}
The logic behind this is that the compatibility penalty is monotonically related to the surface position  difference. In this way, the number of training parameters for the compatibility matrix is reduced from $Z \times Z$ to $1$.

\subsection{Shape-Aware Patch Generation}
\begin{figure}[h]

    \centering
    \includegraphics[width=1.0\linewidth]{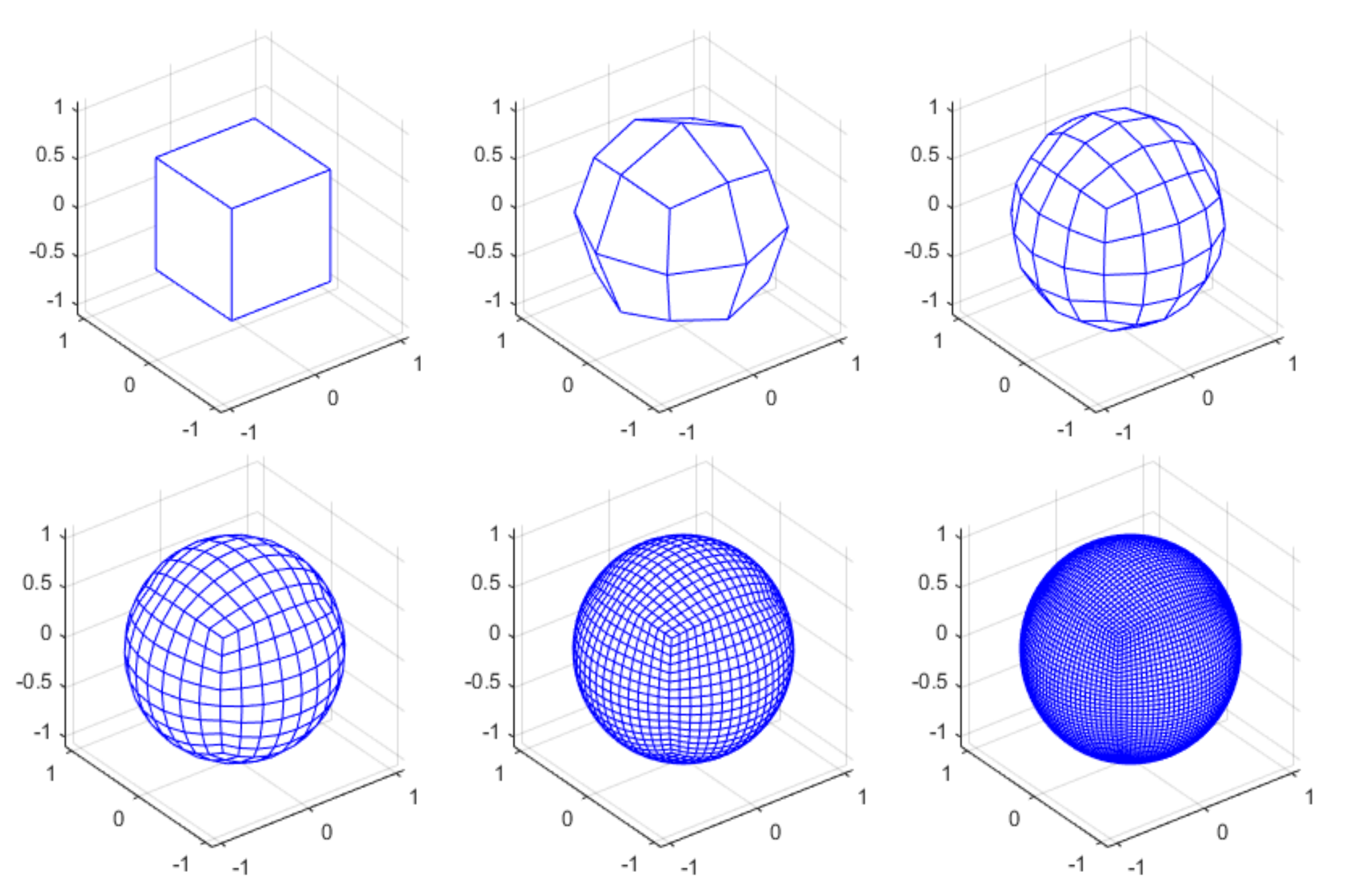}

    \caption{Quadrilateral parameterization of the unit sphere  $S_{quad}$ with the base case and the following 5 recursions, shown from the left to the right and the top to the bottom.}

    \label{fig:sphere_quad}
\end{figure}
For real applications, two obstacles need to be overcome first so that we can model the {\em surface-based} segmentation as terrain-like surfaces segmentation using CNNs. 1) We need to unfold  the surface into a terrain-like surface, on which our {\em surface-based} segmentation is defined. 2) The unfolded image or patch volumes should have a rectangular cuboid grid structure in $3$-D, so that the traditional CNNs can be applied.
\par
In our previous work on  GS segmentation~\cite{li2006optimal,garvin2009automated, yin2010logismos, oguz2014logismos,garvin2008intraretinal, song2013optimal},  we developed effective methods for unfolding the boundary surface of a target object into a terrain-like surface. For a tubular or star-shaped object, it is straightforward to apply a cylindrical or polar coordinate transformation to unfold the target object. For a more complex object, a pre-segmentation is obtained to approximate the (unknown) surface for the target object boundary. A triangulated mesh $\mathcal{P}_{tri}$ is then generated for the pre-segmented surface, which defines the global shape of the target object, including the neighboring relations among voxels on the sought surface. For each vertex of the mesh, a column of voxels is created by resampling the input image along a ray intersecting the vertex (one ray per mesh vertex) based on the medial surface technique~\cite{yin2010logismos}, the electric lines of force theory~\cite{yin2009electric} or gradient vector flows~\cite{oguz2014logismos} to capture the surface location on each column. The adjacency among columns is specified by the mesh $\mathcal{P}_{tri}$. Each sought surface is ensured to cut each column exactly once, thus the boundary surface of the target object is unfolded as a terrain-like surface with respect to the resampled columns. Unfortunately, the resampled image volume with the column structure defined by the triangulated mesh $\mathcal{P}_{tri}$ may not be feasible for us to apply CNNs. The key obstacle is that each column may have variable neighboring columns, thus the whole image volume does not have a regular cuboid grid structure. To resolve that problem, we propose resampling the triangulated mesh $\mathcal{P}_{tri}$ of the pre-segmented surface to form a quadrilateral mesh $\mathcal{P}_{quad}$ by harmonic mapping~\cite{zhang1999harmonic}. The quadrilateral mesh $\mathcal{P}_{quad}$ of the pre-segmented surface is then divided into 6 patches for the training and inference of our proposed segmentation network.
\par
The proposed shape-aware patch generation is detailed, as follows.
\paragraph{Harmonic Mapping}
The triangulated mesh $\mathcal{P}_{tri}$ of the pre-segmented surface, which should be a genus-0 closed surface (otherwise, it needs to close it artificially), is {\em harmonically mapped} to a unit sphere to obtain a triangulated spherical mesh $\mathcal{S}_{tri}$ using the algorithm in \cite{choi2015flash}. 
The harmonic mapping, for a genus-0 closed surface, is conformal ~\cite{schoen1997lectures}, which preserves both angles and orientations, and  is suitable for our application. 
\paragraph{Quadrilateral Parameterization of a Unit Sphere}
The unit sphere can be parameterized by a quadrilateral mesh (except 8 grid points, which only had 3 neighbors), denoted by $S_{quad}$. This parameterization proceeds in a recursive way. The base quadrilateral mesh  is an  inscribed cube of the unit sphere. In every recursion, each face (a square) is divided equally into four squares, i.e. the middle point of each edge and the center of each face are moved outwards exactly to the unit sphere. This process is demonstrated in Fig.~\ref{fig:sphere_quad}. This produces a higher grid resolution on the surface with more recursive iterations.
In our experiments, the number of recursions is chosen as 5. In other words, for each face of the base quadrilateral mesh, the number of quadrilateral faces in the mesh  increases from the base $1\times1$ to $(2^5) \times (2^5)$, which corresponds to a square grid with a size $(2^5+1) \times (2^5+1)$.
\begin{figure}

    \centering
        \begin{tabular}{c c }
		    \includegraphics[width=0.4\linewidth, height=0.4\linewidth]{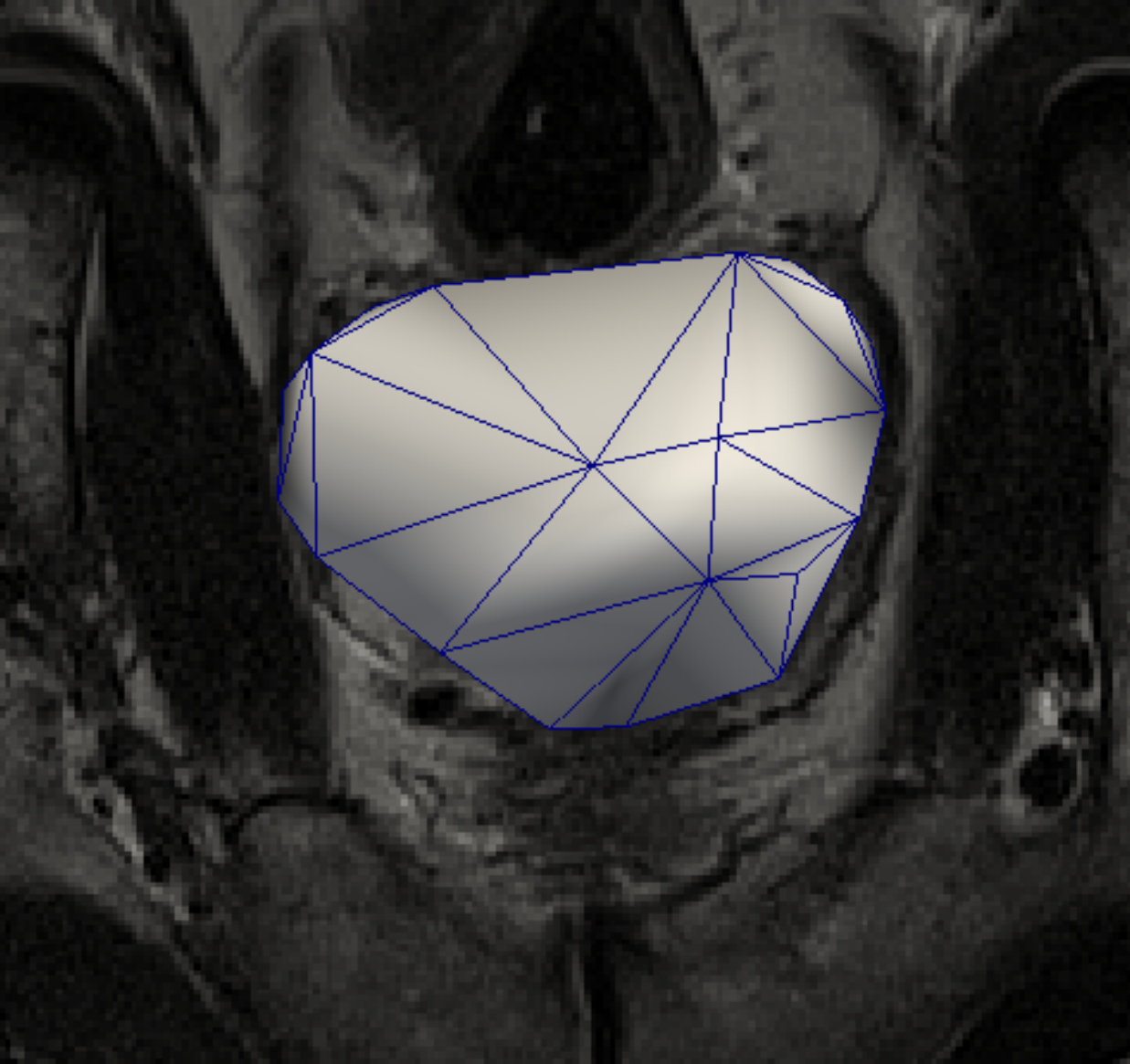} &
		    \includegraphics[width=0.4\linewidth, height=0.4\linewidth]{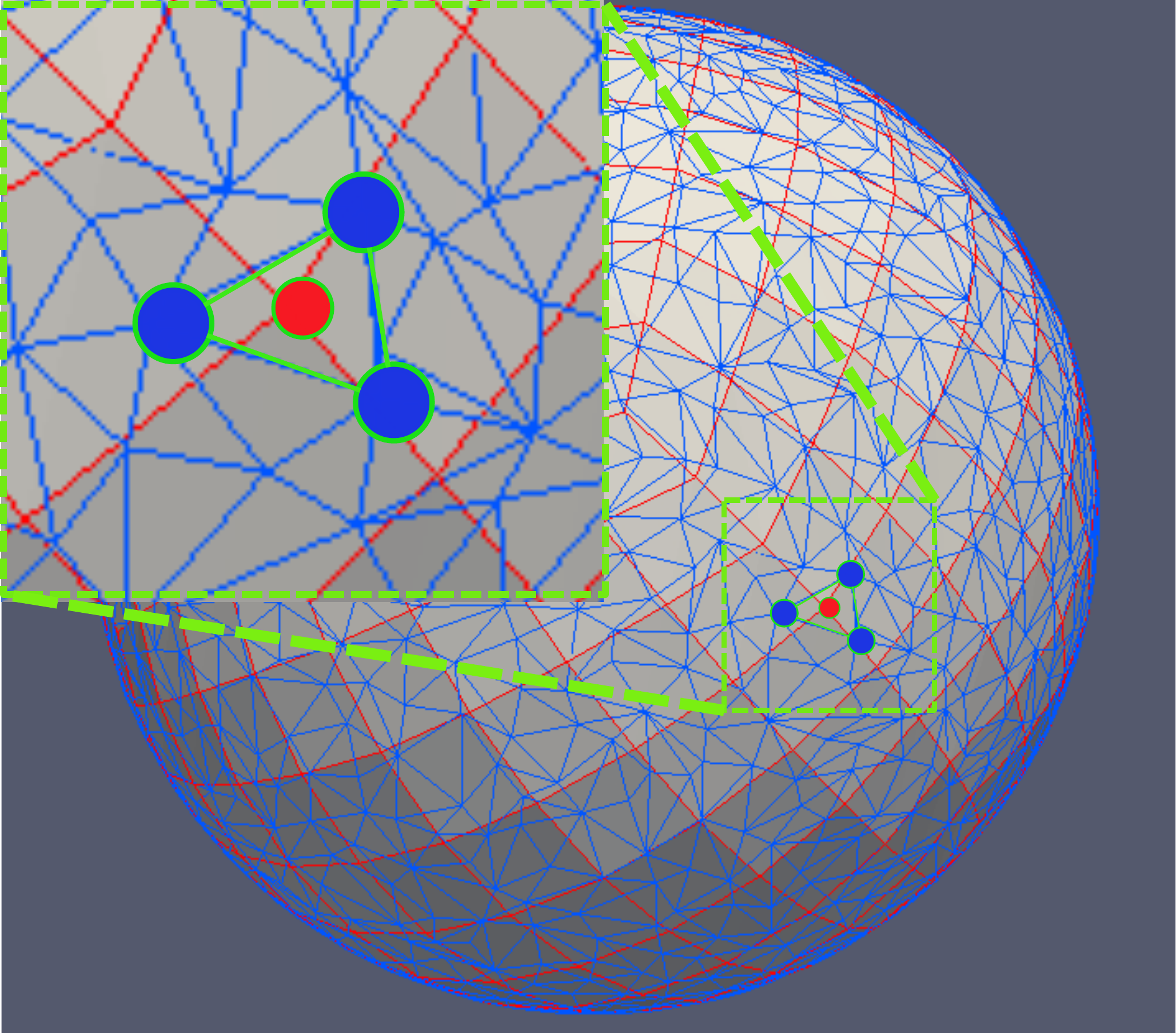} 
		   \\
		 (a) & (b)\\
		    \includegraphics[width=0.4\linewidth, height=0.4\linewidth]{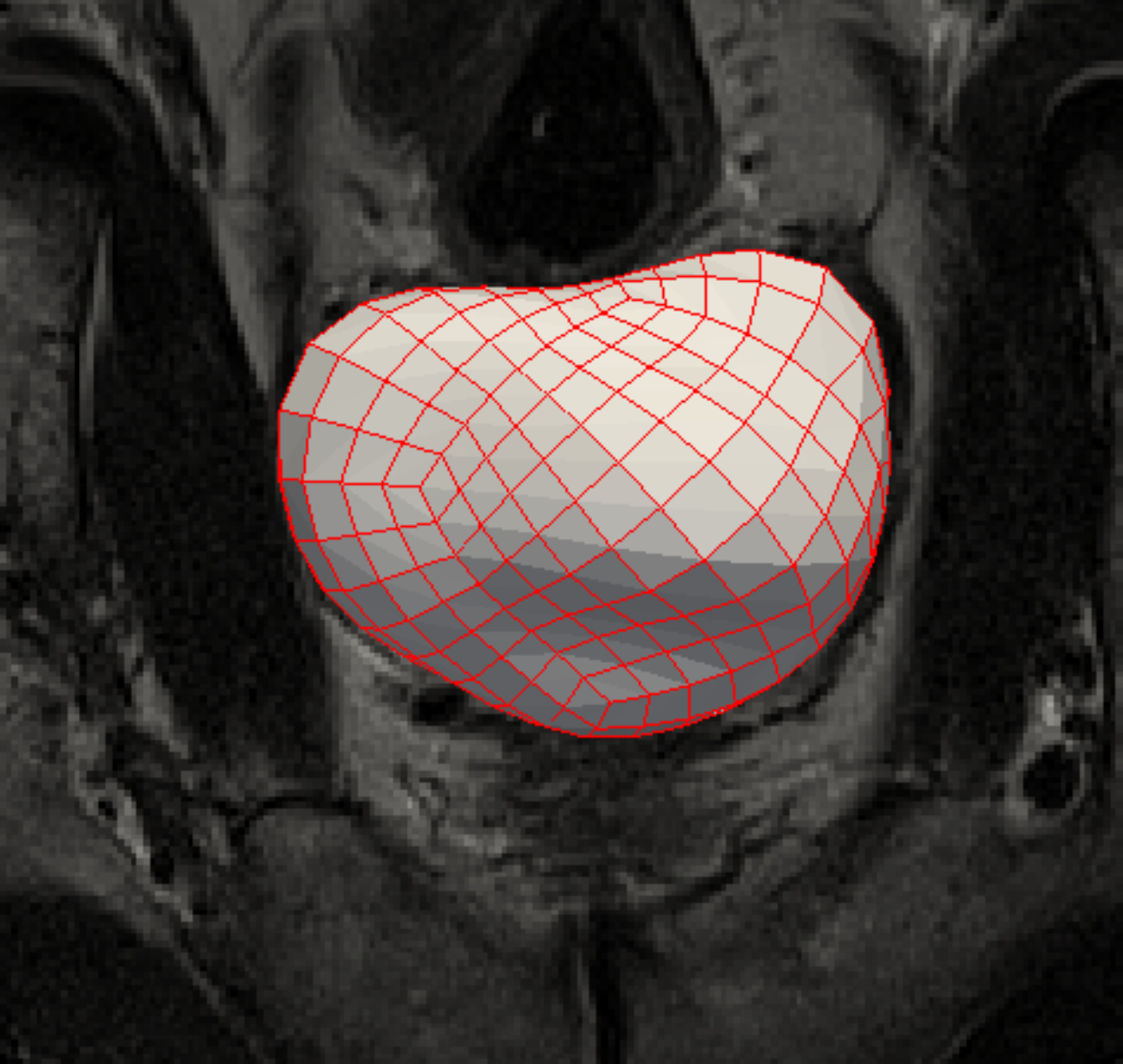} &
		    \includegraphics[width=0.4\linewidth, height=0.4\linewidth]{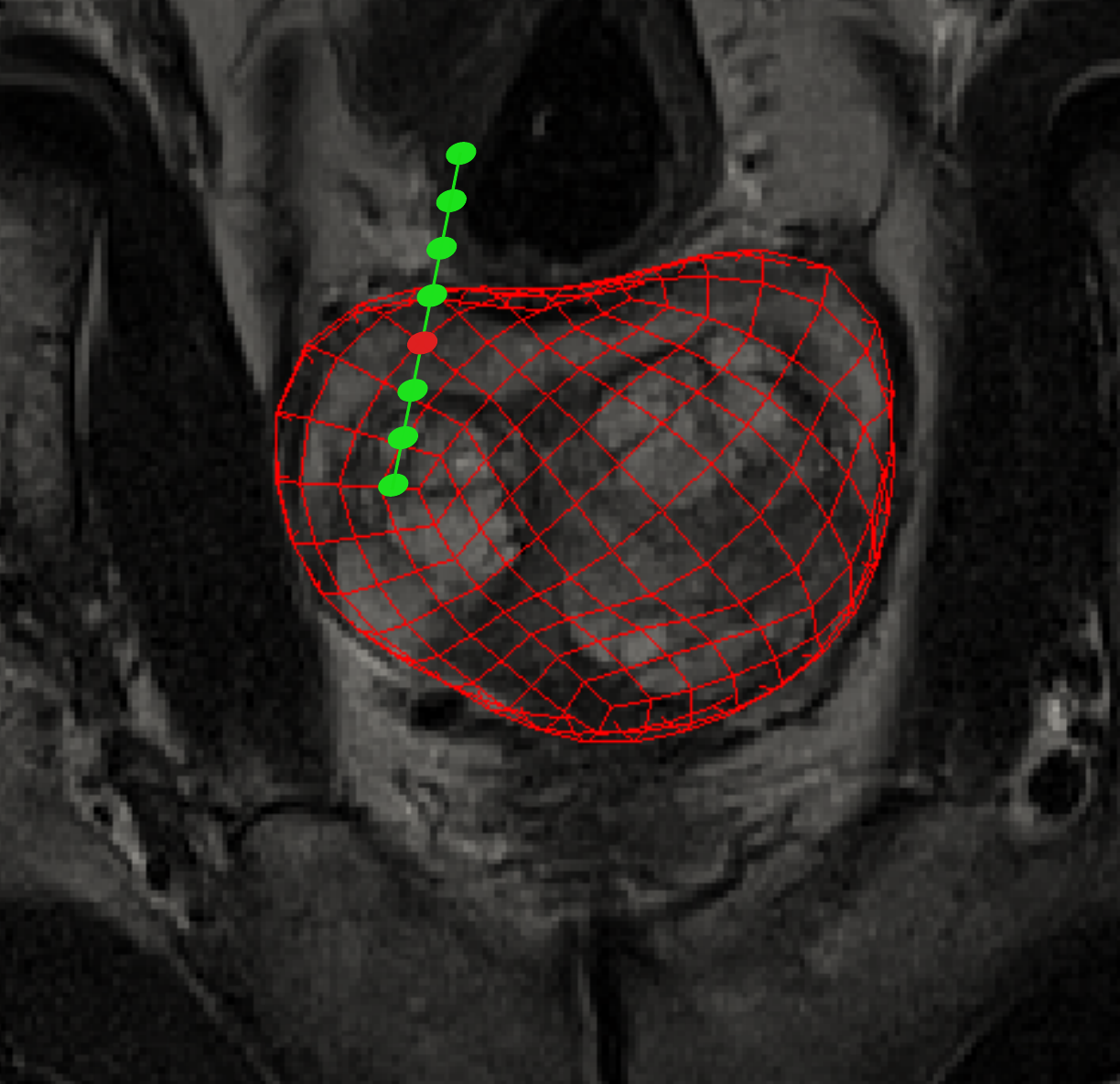}
		\\
		    (c) & (d) 
	    \end{tabular}

    \caption{The proposed process of patch generation. (a) Triangular pre-segmentation mesh $\mathcal{P}_{tri}$ (blue). Downsampling was done for demonstration. (b) Overlay of the harmonically mapped {\em tri preseg}  surface $\mathcal{S}_{tri}$ (blue) and the {\em quad} parameterized unit sphere $\mathcal{S}_{quad}$ (red). For the red vertex $v^{(s)}_{quad}$ on $\mathcal{S}_{quad}$, the corresponding face $f^{(s)}_{tri}$ is demonstrated as the triangle with three vertices colored as blue and edges colored as green. (c) {\em Quad preseg} mesh surface $\mathcal{P}_{quad}$. (d) Sampling a column in normal direction for each $v^{(p)}_{quad} \in \mathcal{P}_{quad}$ as its feature.}

    \label{fig:quad_overlay}
\end{figure}
\paragraph{Quadrilateral Mesh of the Pre-segmented Surface}
Recall that  $\mathcal{S}_{tri}$ is a triangular mesh obtained by projecting the triangulated mesh $\mathcal{P}_{tri}$ of the pre-segmented surface to a unit sphere via harmonic mapping. As $\mathcal{S}_{quad}$ is a quadrilateral mesh of the same sphere, both $\mathcal{S}_{tri}$ and $\mathcal{S}_{quad}$ are defined in the same manifold space, thus can be overlaid on each other.
For each vertex $v^{(s)}_{quad} \in \mathcal{S}_{quad}$, the corresponding triangular face $f^{(s)}_{tri}$ of  $\mathcal{S}_{tri}$ with $v^{(s)}_{quad} \in f^{(s)}_{tri}$ can be found. The Barycentric coordinate of this vertex $v^{(s)}_{quad}$ can then be computed with respect to the triangle $f^{(s)}_{tri}$, which is a triple of numbers $(t_1, t_2, t_3)$ corresponding to masses placed at the vertices of $f^{(s)}_{tri}$ such that $v^{(s)}_{quad}$ is the geometric centroid of $f^{(s)}_{tri}$ with the three masses, i.e. $v^{(s)}_{quad}=t_1 v_1+t_2 v_2+t_3 v_3$, where $v_1, v_2$ and $v_3$ are the three vertices of the triangle face $f^{(s)}_{tri}$. It is apparent that the new parameterization of $v^{(s)}_{quad} \in \mathcal{S}_{quad}$ with respect to $\mathcal{S}_{tri}$, i.e. $v^{(s)}_{quad}\rightarrow (f^{(s)}_{tri}, t_1, t_2, t_3)$, is bijective, since $\mathcal{S}_{quad}$ is the same manifold to $\mathcal{S}_{tri}$. As $\mathcal{S}_{tri}$ is a harmonic map of $\mathcal{P}_{tri}$, the faces $f^{(p)}_{tri} \in \mathcal{P}_{tri}$ and the ones $f^{(s)}_{tri} \in \mathcal{S}_{tri}$ have a one-to-one correspondence. Thus, each  $v^{(s)}_{quad} \in S_{quad}$ can be mapped to its corresponding point $(f^{(p)}_{tri}, t_1, t_2, t_3)$ in $\mathcal{P}_{tri}$. In other words, for each $v^{(s)}_{quad}$, we can get its unique corresponding point  on the original triangulated pre-segmentation surface $\mathcal{P}_{tri}$. Recall that the harmonic mapping preserves angels and orientations, and then the local ordering information.
In this way, a guaranteed quadrilateral  mesh (except 8 vertices)  for the triangulated pre-segmentation surface can be realized, which is denoted by $\mathcal{P}_{quad}$.
This quadrilateral remeshing process works for all genus-0 closed surfaces, which is illustrated in Fig.~\ref{fig:quad_overlay} (b-c).

\paragraph{Sampling Columns to Generate Patches}
After the quadrilateral remeshing, for each vertex $v \in \mathcal{P}_{quad}$, we sample a column of voxels with a certain length and resolution in the normal direction, which is treated as the image feature for that  vertex and corresponds to one column in our problem definition (Fig.~\ref{fig:quad_overlay} (d)).
The quadrilateral surface mesh $\mathcal{P}_{quad}$ is a $2$-D manifold. Thus, after extending in the image feature column dimension, a $3$-D image volume $\mathcal{I}$ is generated. 
In addition, the unit sphere with our quadrilateral parameterization can be easily split  into 6 pieces, which correspond to the 6 faces of the initially inscribed cube. The quadrilateral mesh $\mathcal{P}_{quad}$ can then be decomposed into 6 corresponding pieces as well. Each piece of $\mathcal{P}_{quad}$ defines a $3$-D image patch $\mathcal{I}$, in which the target surface is a terrain-like one.

\paragraph{Ground Truth Generation}
When the $\mathcal{P}_{quad}$ is derived, the truth voxel on the target surface for each vertex $v \in \mathcal{P}_{quad}$ and the corresponding column can be defined as the nearest neighbor voxel to the intersection of the manual segmentation mesh with the normal of $\mathcal{P}_{quad}$ at $v$. One should note that the ground truth position is defined to be relative to the pre-segmented quadrilateral surface $\mathcal{P}_{quad}$.
\subsection{Network Architecture}

\begin{figure}[h!]
    \centering
	\includegraphics[width=1.0\linewidth, height=0.75\linewidth]{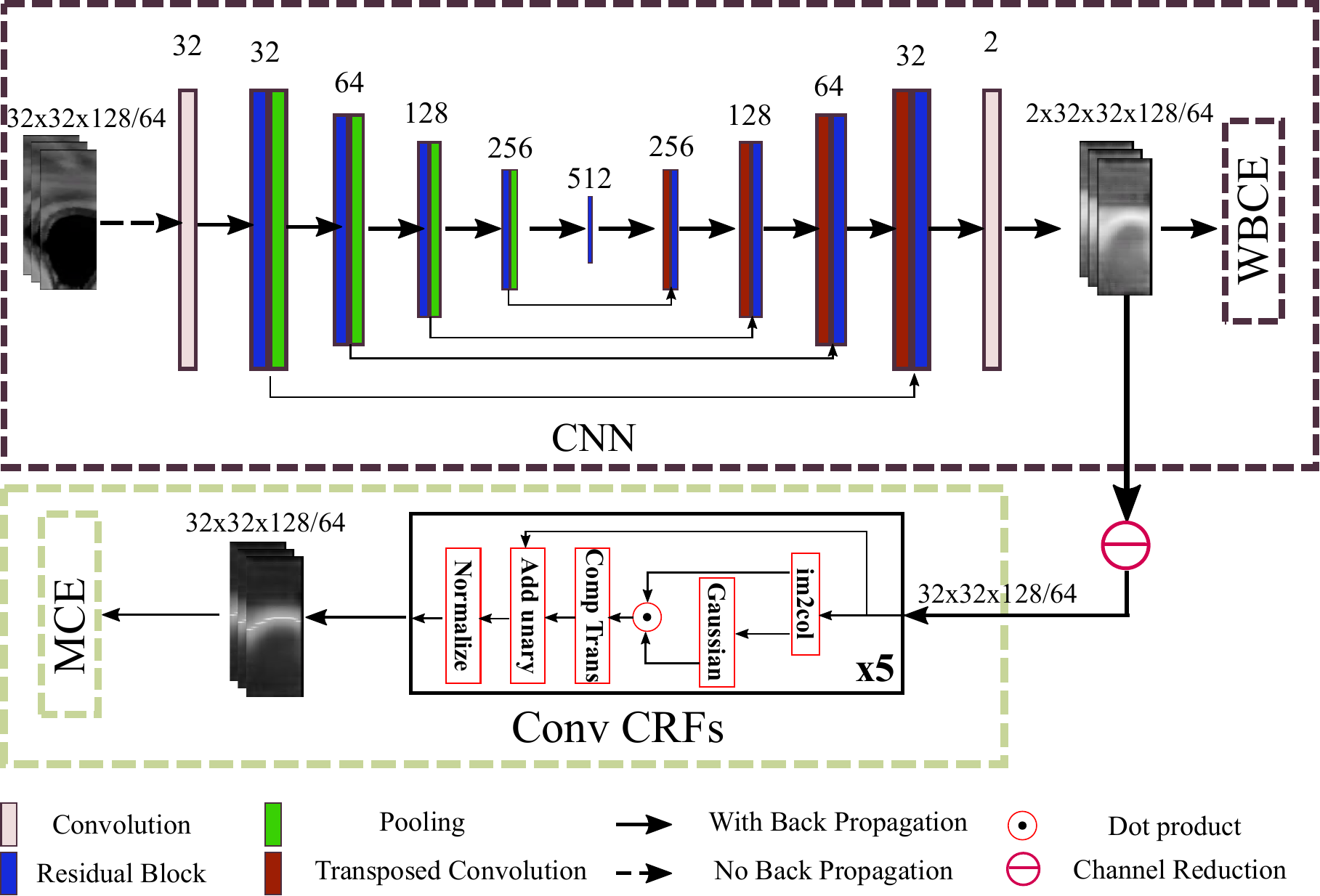} 
		   
    \caption{The proposed surface segmentation network. It consists of the  proposed CNN and the   proposed CRFs layer, from the top to the bottom. For convolution and residual blocks, the number indicates dimensions of feature maps, and $3 \times 3 \times 3$ convolution kernels are utilized except the last convolution layer, where $1 \times 1 \times 1$ kernels are chosen; for pooling, $3 \times 3 \times 3$ convolution kernels with strides 2 are utilized; for transposed convolution layers, $2 \times 2 \times 2$ kernels and strides 2 are utilized. The {\em Channel Reduction} function is applied to subtract being {\em non-surface} logit from being {\em surface} logit for each voxel and we treat the difference as the logit of being {\em surface} within each column, and we select the voxel with the biggest being {\em surface} logit as the surface prediction.}

    \label{fig:architecture}
\end{figure}
\noindent
The network is designed for direct surface segmentation.
The architecture consists of two main parts: a $3$-D encoder-decoder CNN for surface probability map generation, and a trainable CRFs for modeling the unary and pairwise terms simultaneously,
as demonstrated in Fig.~\ref{fig:architecture}. 
\subsubsection{The $3$-D encoder-decoder CNN}
In medical image segmentation, the encoder-decoder CNNs are widely used. We adopt a similar architecture to generate the surface probability map, i.e. the unary term in our surface model (Eqn.~(\ref{surf-model})). 
 Global skip connections are built as in \cite{ronneberger2015u}, as well as short or local skip connections as utilized in He \textit{et al.} \cite{he2016deep}, where a unit block is called a {\em residual} block. Those connections are used to mitigate the gradient vanishing problem. As output of  the encoder-decoder CNN, a two-channel probability map for the target surface is generated. During the pre-training of the CNN, the supervision is added in  with a  {\em weighted binary cross-entropy} (WBCE) loss. The ground truth here is a binary mask of the same size as the input patch. The key difference here from those in {\em region-based} segmentation neural networks, is that 1 and 0 represent the target surface and background, respectively. Thus, the resulting classification problem is highly imbalanced. We introduce the WBCE loss to alleviate the problem.
\subsubsection{The CRFs network}
CRFs are introduced to explicitly model the unary and pairwise terms simultaneously. CRFs are more commonly associated with {\em region-based} segmentation networks making this one of the first occasions where it is applied to the {\em surface-based} segmentation. In Shah \textit{et al.}'s method, a FC layer was utilized to directly regress the surface position but from the feature maps in a {\em low} spatial resolution. 
\par
The fully-connected CRFs model was first introduced to semantic segmentation by Kr{\"a}henb{\"u}hl and Koltun~\cite{krahenbuhl2011efficient}, which is known as DenseCRF. Although DenseCRF utilized a mean-field approximation inference, it achieved significantly improved results with an efficient inference. This has become the backbone for most CRFs models. The mean-field inference of a DenseCRF model can be incorporated into a neural network, which was developed by Zheng \textit{et al.} \cite{zheng2015conditional}. This enables the joint training of CNNs and CRFs by simple back propagation and was named CRF-as-RNN. In CRF-as-RNN, the message-passing step is the bottleneck. The exact computation is quadratic in the number of pixels, and therefore is not efficient for $3$-D image applications. To alleviate this issue, a permutohedral lattice approximation was utilized. However, computing it efficiently on GPU is non-trivial or intractable to realize. In addition, an efficient gradient computation of the permutohedral lattice approximation, is also a non-trivial problem. This may hinder the learning of some parameters, e.g. $\theta_{1}$, $\theta_{2}$, and $\theta_{3}$ in the kernel term Eqn.~(\ref{eqn-kernel}). In the convolutional CRFs \cite{teichmann2018convolutional}, the message passing is reformulated to be a convolution with a truncated Gaussian kernel and can be implemented in a similar way to the regular convolutions in CNNs. Therefore, the convolutional CRFs are utilized in the proposed method.

\subsubsection{The loss functions}
The {\em cross-entropy} (CE) loss is utilized both for the pre-training of the CNN and the fine tuning of the CNN+CRFs network. For pre-training, it is a {\em binary cross-entropy} (BCE) loss, since the encoder-decoder is meant to output the probabilities of each voxel being on the surface or not. Also, as the number of voxels on the surface is normally much less than that of those not on the surface, a {\em weighted binary cross-entropy} (WBCE) loss is used. For the CNN+CRFs  fine tuning, the problem is modeled as a multinomial classification and therefore a {\em multinomial cross-entropy} (MCE) loss is chosen. The fine tuning is in an end-to-end fashion.  
The loss functions and training strategies for the proposed method are summarized in Table. \ref{table:loss}.
\begin{table}

	\centering
	\caption{Losses  and training strategies for the proposed segmentation methods.}

	\begin{tabular}{|>{\centering\arraybackslash}m{0.15\linewidth}|>{\centering\arraybackslash}m{0.2\linewidth}|>{\centering\arraybackslash}m{0.15\linewidth}|>{\centering\arraybackslash}m{0.25\linewidth}|} 
		\hline
		  & CNN Loss & CRFs Loss & Training Strategy \\ 
		\hline
		proposed CNN & WBCE & - & - \\
		 \hline
		proposed CNN+CRFs & - & MCE & Pretrain CNN and then fine tune CNN+CRFs \\
		  
		  \hline
	\end{tabular}
	\label{table:loss}

\end{table}
In the following two sections, the proposed method was applied to the prostate MRI segmentation and the spleen CT segmentation.
\section{Application to the Prostate MRI Segmentation}
\subsection{Experiment Design}
\subsubsection{Data}
The dataset is provided by the NCI-ISBI 2013 Challenge - Automated Segmentation of Prostate Structures \cite{bloch2015nci}. This dataset has two labels: peripheral zone (PZ) and central gland (CG). We treat both of them as prostate, since the single surface segmentation is considered in this work. The challenge data set consists of the training set (60 cases), the leader board set (10 cases) and the test set (10 cases). As the challenge is closed, only the training and leader board data with annotation (70 cases in total) were used for our experiments. 10-fold cross validation was applied on that dataset. For each fold, the training, validation and test sets consist of 50, 10 and 10 cases, respectively.

\subsubsection{Pre-segmentation}
Our method needs to pre-segment the target object to obtain its basic shape. The $3$-D patches are then generated based on the quadrilateral mesh of the pre-segmented surface.
To test the robustness of the proposed method to pre-segmentation and the column length (the resolution is fixed), two pre-segmentation methods were explored. The first method was to fit a fixed size ellipsoid to the input image. The second method was to coarsely fit a mean shape to the user defined bounding box, which produces a more accurate pre-segmentation.  With a better pre-segmentation, we can sample the feature columns in a shorter length. All volumes were resampled to be isotropic with voxel resolution of $0.625\times 0.625 \times 0.625$ mm$^3$ and normalized to have a zero mean and a unit variance.

\paragraph{Ellipsoid Pre-segmentation}
For simplicity, an ellipsoid with three principal semi-axes of length $25$ mm, $22$ mm and $25$ mm, was used for the pre-segmentation. The centers of the ellipsoids were picked by users. As this pre-segmentation is far from perfect (The average Dice similarity coefficient (DSC) was around 0.7), longer columns should be sampled to ensure the voxels on the target surface be included. The column length for the ellipsoid pre-segmentation  was set to be 128 and the resolution was $0.625$ mm. 

\paragraph{Mean Shape Pre-segmentation}
For training data, we aligned all images to one randomly picked reference image based on the centers of the manually segmented prostates, such that all target objects in the training set were coarsely aligned. And then zero level set of average surface distance maps would be the mean shape. For the test data, based on the bounding box manually defined, we fitted the mean shape into the bounding box by only changing the value of the level set, i.e. the mean shape was only allowed to do the scaling transformation.
The column length under this setting was reduced to 64 (resolution=$0.625$ mm) as the pre-segmentation was more accurate (the average DSC was about 0.78).

\subsubsection{Data Augmentation}\label{sec:aug}
Rotations (with degrees of $90^\circ$, $180^\circ$, and $270^\circ$), flippings in two in-plane directions, combination of both rotation and flipping, as well as simple random translations in the $z$ direction, were applied. In total, the amount of the training patch was enlarged by a factor of 14, from $50\times6$ to $50\times6\times14$.

\subsubsection{Hyper Parameters}
The proposed network was implemented with Pytorch \cite{paszke2017automatic}. The network was initialized with  Xavier normal initialization \cite{glorot2010understanding}. The patch size ($X\times Y \times Z$) was $32\times32\times128$ or $32\times32\times64$ for two different pre-segmentation settings, in which $32\times32$ represents the in-plane size (i.e., the number of columns in each patch), and the resolution on the column direction was $0.625$ mm. The training of the proposed CNN+CRFs network consists of two stages: 1) pre-training the CNN network and 2) fine-tuning the whole CNN+CRFs network.

\paragraph{Pre-training the proposed CNN network}
Adam optimizer \cite{kingma2014adam} with learning rate $10^{-3}$, was chosen for the pre-training. We let it run for 50 epochs. The weight for the WBCE loss was the column length in the patch, which was about inversely proportional to the ratio between the number of voxels on the surface and that of non-surface voxels in the ground truth annotation. 

\paragraph{Fine-tuning the CNN+CRFs network}
The fine-tuning of the whole CNN+CRFs network was done in an end-to-end fashion. During the fine tuning, the learning rate of Adam was $10^{-5}$, and the training ran for 50 epochs. Only the MCE loss function following the CRFs layer was utilized. The initialization of parameters in the CRFs layer is detailed in Table. \ref{table:crf_para}.
\begin{table}[h]
	\centering
	\caption{Initialization of parameters in the CRFs layer for the prostate MRI segmentation.}

	\begin{tabular}{|c|c|c|c|c|c|} 
		\hline
		 $w_p$ & $ w_1$ & $\theta_{1}$  & $\theta_{2}$ & $\theta_{3}$ & $\theta_{\text{comp}}$\\ 
		\hline
		1 &  3 & 5 & 0.2  & 5 & 5\\
		\hline
	\end{tabular}
	\label{table:crf_para}

\end{table}

\subsection{Evaluation Metrics}
Three metrics -- Dice similarity coefficient (DSC), Hausdorff distance (HD) (the greatest of all the distances between each point on the computed surface and its closest point on the reference surface), and the average surface distance (ASD) (the average over the shortest distances between the points on the computed surface and the reference surface), were engaged to evaluate results of segmentation. The DSC is defined, as follows.
\begin{equation}
DSC = \frac{2|V_g \cap V_p|}{|V_g| + |V_p|},
\end{equation}
where $|V_g|$ is the number of voxels of prostate in the ground truth, $|V_p|$ is the number of prostate voxels in the prediction, and $|V_g \cap V_p|$ is the number of overlapping prostate voxels between the ground truth and the prediction. 

The HD between the two surfaces $S_1$ and $S_2$ is computed, with
\begin{equation}
HD(S_1, S_2) = \text{max} \{ \underset{s_1 \in S_1}{\text{max}} d(s_1, S_2),  \underset{s_2 \in S_2}{\text{max}} d(s_2, S_1)\}.
\end{equation}
where $d(s, S)$ is the distance from a voxel $s$  to a surface $S$, which is defined as:
\begin{equation}
d(s, S) = \underset{s' \in S}{\text{min}} ||s - s'||.
\end{equation}

The ASD is defined as:
\begin{equation}
\begin{split}
ASD(S_1, S_2) = \frac{1}{|S_1|+|S_2|} & \big(\underset{s_1 \in S_1}{\sum} d(s_1, S_2)  \\
&  + \underset{s_2 \in S_2}{\sum} d(s_2, S_1)\big),
\end{split}
\end{equation}
where $|S_1|$ and $|S_2|$ are the number of voxels in surface $S_1$ and $S_2$, respectively.

\subsection{Results}
The quantitative segmentation results of different methods are listed in Table. \ref{table:overall}. 
In Table. \ref{table:overall}, our results were derived using only NCI-ISBI data, while the compared methods, FCN~\cite{long2015fully}, V-net~\cite{milletari2016v}, U-net~\cite{ronneberger2015u}, and PSNet~\cite{tian2018psnet} made use of additional in-house data and the Promise12 data \cite{litjens2014evaluation} for their network training. For all other methods, only NCI-ISBI dataset was used. In other words, the results of the first four methods were derived using around double the number of training cases and a similar number of validation and test cases. 
With respect to  DSC metric, our method outperformed FCN, V-net, U-net and PSNet, and was comparable to the deep learning state-of-the-art GCA-Net~\cite{jia20183d} and another state-of-the-art traditional method~\cite{tian2017supervoxel} which combined the supervoxel method, Graph Cut and Active Contour Model (ACM). With respect to the surface distance related metrics (i.e., HD and ASD), the proposed method significantly outperformed all the compared methods. Compared to most of the {\em region-based} deep learning methods, such as FCN, V-net, and U-net,  our proposed method does not need any post processing, e.g. morphological operations, to remove holes within the segmented object. The GS method shares the same merit as the proposed method. However, due to the need to manually design the cost function (i.e., to generate the probability maps), even the solution to the energy minimization problem is global optimal, its performance was inferior to the proposed method and other deep learning based methods.

 \begin{table}[h]

	\centering
	\caption{Quantitative comparison with other prostate segmentation methods. The first four methods used both the NCI-ISBI dataset and the Promise12 dataset for training. The remaining methods including our method only used the NCI-ISBI dataset.}

	\begin{tabular}{|c|c|c|c|} 
		\hline
		 & DSC & ASD (mm) & HD (mm) \\ 
	
		\hline
		FCN \cite{long2015fully} & 0.79$\pm$0.06 & 4.8$\pm$1.1 & 11.9$\pm$4.8  \\
		\hline
		V-net \cite{milletari2016v} & 0.83$\pm$0.05 & 3.4$\pm$1.2 & 9.5$\pm$3.9  \\
		\hline
		U-net \cite{ronneberger2015u} & 0.84$\pm$0.05 & 3.3$\pm$1.0 & 10.1$\pm$3.2  \\
		\hline
		PSNet \cite{tian2018psnet} & 0.85$\pm$0.04 & 3.0$\pm$0.9 & 9.3$\pm$3.5  \\
		\hline
		GS  & 0.80$\pm$0.04 & 2.7$\pm$0.6 & 13.9$\pm$1.8\\
		\hline
		SupervoxelGraphCutACM \cite{tian2017supervoxel} & \textbf{0.88$\pm$0.02} & - & -\\
		\hline
		
		GCA-Net \cite{jia20183d} & \textbf{0.88} & 2.2 & -\\
		\hline
		proposed CNN+CRFs & \textbf{0.88$\pm$0.03} & \textbf{1.4$\pm$0.3} & \textbf{8.2$\pm$3.6} \\
		\hline
	\end{tabular}
	\label{table:overall}

\end{table}
\subsection{Robustness to Different Pre-segmentations}
The results with two different pre-segmentations are shown in  Table. \ref{table:preseg}. Better pre-segmentations and shorter image columns improved the DSC and HD performance consistently. The ASDs  were comparable. The results basically indicated that although better pre-segmentations could help, our method was not sensitive to different pre-segmentations as long as  the basic topology of the target surface was correct. 

\begin{table}[h]

	\centering
	\caption{Prostate segmentation results of the proposed methods with different pre-segmentations and column lengths.}

	\begin{tabular}{|>{\centering\arraybackslash}m{0.15\linewidth}|>{\centering\arraybackslash}m{0.15\linewidth}|>{\centering\arraybackslash}m{0.12\linewidth}|>{\centering\arraybackslash}m{0.12\linewidth}|>{\centering\arraybackslash}m{0.12\linewidth}|} 
		\hline
		 Preseg, Col length &  & DSC & ASD (mm) & HD (mm) \\ 
		\hline
		Ellipsoid, 128 & proposed CNN+CRFs  & 0.86$\pm$0.05 & \textbf{1.4$\pm	$0.5} & 	9.6$\pm$5.2 \\

		\hline
		
		Mean shape, 64 & proposed CNN+CRFs & \textbf{0.88$\pm$0.03} & \textbf{1.4$\pm$0.3} & \textbf{8.2$\pm$3.6} \\
		
		\hline
	\end{tabular}
	\label{table:preseg}

\end{table}
\subsection{Ablation Study}
\begin{figure}[h]

    \centering
        \begin{tabular}{c c }
		    \includegraphics[width=0.4\linewidth, height=0.4\linewidth]{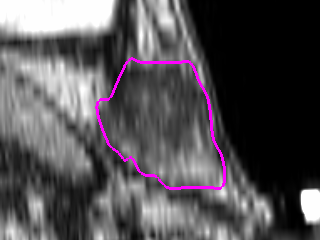} &
		    \includegraphics[width=0.4\linewidth, height=0.4\linewidth]{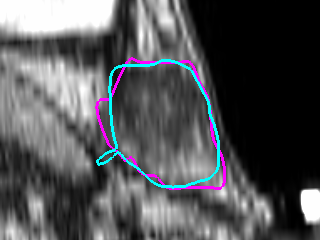} 
		   \\
		 (a) & (b)\\
		    \includegraphics[width=0.4\linewidth, height=0.4\linewidth]{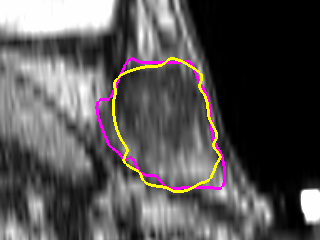} &
		    \includegraphics[width=0.4\linewidth, height=0.4\linewidth]{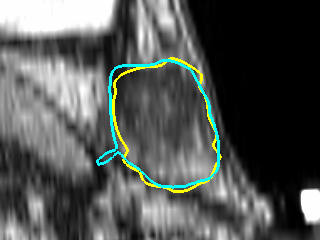}
		\\
		    (c) & (d) \\
	    \end{tabular}

    \caption{Sample prostate segmentation results of proposed methods in the sagittal view. (a) The original image overlaid with the ground truth (purple). (b) The result of the proposed CNN (light blue).  ASD and HD were $1.21$ mm and $13.8$ mm. (c)The result of the proposed CNN+CRFs (yellow). ASD and HD were improved to $1.20$ mm and $6.0$ mm. (d) Overlay of the result of CNN and that of the CNN+CRFs.  }

    \label{fig:ablation}
\end{figure}

We also investigated the ablation study to verify if the CRFs layer could improve the surface segmentation. The proposed CNN method was used to directly infer the segmentation results and was compared to the CNN+CRFs model. The ablation study results are shown in Table. \ref{table:ablation}. Although the CRFs layer did not improve the DSC performance, it did improve the segmentation performance according to the surface related metrics. One sample of the improvement is illustrated in Fig.~\ref{fig:ablation}.
\begin{table}[h]

	\centering
	\caption{Prostate segmentation results with or without the CRFs.}

	\begin{tabular}{|>{\centering\arraybackslash}m{0.15\linewidth}|>{\centering\arraybackslash}m{0.15\linewidth}|>{\centering\arraybackslash}m{0.12\linewidth}|>{\centering\arraybackslash}m{0.12\linewidth}|>{\centering\arraybackslash}m{0.12\linewidth}|} 
		\hline
		 Preseg, Col length &  & DSC & ASD (mm) & HD (mm) \\ 
		\hline
		 \multirow{2}{4em}{Ellipsoid, 128} & proposed CNN  & \textbf{0.86$\pm$0.05} & 1.5$\pm$0.5 & 11.3$\pm$5.9 \\
		
		& proposed CNN+CRFs  & \textbf{0.86$\pm$0.05} & \textbf{1.4$\pm$0.5} & 	\textbf{9.6$\pm$5.2} \\
		\hline
		
		  \multirow{2}{4em}{Mean shape, 64} & proposed CNN & \textbf{0.88$\pm$0.03} & \textbf{1.4$\pm$0.3} & 8.3$\pm$2.9 \\
		& proposed CNN+CRFs & \textbf{0.88$\pm$0.03} & \textbf{1.4$\pm$0.3} & \textbf{8.2$\pm$3.6} \\
		\hline
	\end{tabular}
	\label{table:ablation}

\end{table}

\section{Application to the Spleen CT Segmentation}
\subsection{Experiment Design}
\subsubsection{Data}
The dataset is provided by Task09 of Medical Segmentation Decathlon (MSD) challenge~\footnote{https://decathlon.grand-challenge.org/Home/}. Only training sets with annotation were utilized. There are 41 cases in total.
 All experiments were conducted with 4-fold cross validation.
 
\subsubsection{Patch Generation}
All volumes were resampled to be isotropic with voxel resolution $0.85\times 0.85 \times 0.85$ mm$^3$ and normalized to have a zero mean and a unit variance. 
For simplicity, a $3$-D V-net was trained as the baseline model. A $3$-D active contour model~\cite{gao20123d} was utilized to provide a coarse segmentation, which was then smoothed to form our pre-segmentation. The generated patch has a size $32\times32\times64$.

\subsubsection{Data Augmentation}
The same augmentation strategy as in Section~\ref{sec:aug} was applied to this task. In total, the number of  training patches was $26\times6\times14$.

\subsubsection{Hyper Parameters}
The network was initialized with  the Xavier normal initialization. The patch size ($X\times Y \times Z$) is $32\times32\times64$, in which $32\times32$ represents the in-plane size (i.e., the number of columns in each patch), and the resolution on the column direction was $0.85$ mm.
\paragraph{V-net}
A public implementation~\footnote{https://github.com/mattmacy/vnet.pytorch} of V-net was used. The patch size is $128\times128\times64$. The BCE was chosen as the loss function. The network was trained using Adam optimizer with a learning rate of $10^{-3}$ for $300$ epochs. The post-processing was applied to remove small regions.

\paragraph{The Proposed CNN+CRFs}
To pre-train the CNN part, Adam optimizer with learning rate of $10^{-4}$ was chosen and it ran for 200 epochs. The weights within the WBCE loss function were $1$ and $64$. 
During the fine tuning, the  MCE loss  was utilized. The learning rate of Adam was $10^{-5}$,  and the training ran for 100 epochs. The initialization of parameters in the CRFs layer is detailed in Table. \ref{table:crf_para_spleen}.
\begin{table}[h]
	\centering
	\caption{Initialization of parameters in CRFs layer for the spleen CT segmentation.}

	\begin{tabular}{|c|c|c|c|c|c|} 
		\hline
		$w_p$ & $w_1$ & $\theta_{1}$  & $\theta_{2}$ & $\theta_{3}$ & $\theta_{\text{comp}}$\\ 
		\hline
		0.3 &  0.2 & 5 & 0.2  & 5 & 5\\
		\hline
	\end{tabular}
	\label{table:crf_para_spleen}

\end{table}

\subsection{Evaluation Metrics}
The DSC, ASD and HD were used to quantify the segmentation results.

\subsection{Performance Comparison}

\begin{figure}[h]

    \centering
        \begin{tabular}{c c }
		    \includegraphics[width=0.4\linewidth, height=0.4\linewidth]{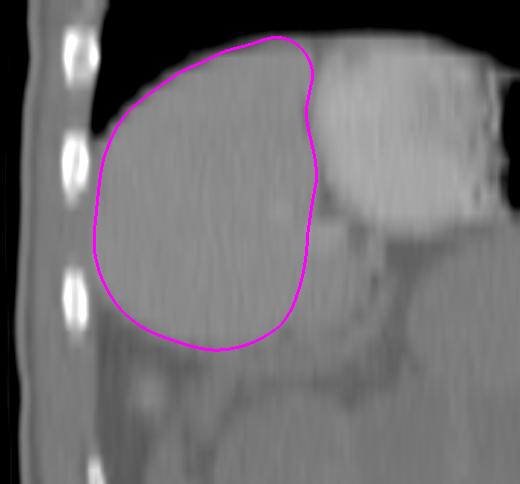} &
		    \includegraphics[width=0.4\linewidth, height=0.4\linewidth]{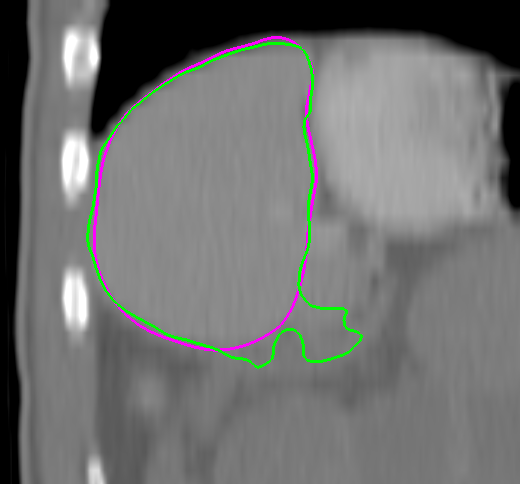} 
		   \\
		 (a) & (b)\\
		    \includegraphics[width=0.4\linewidth, height=0.4\linewidth]{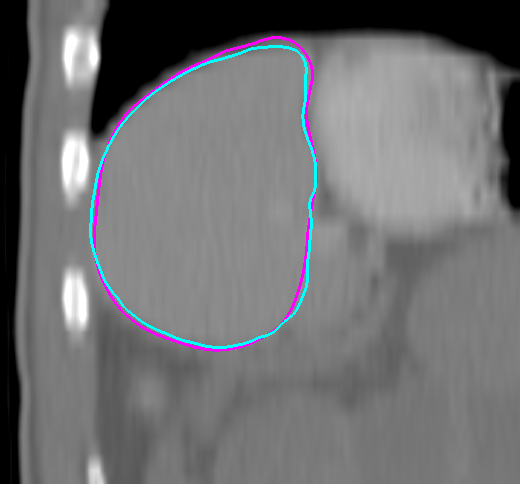} &
		    \includegraphics[width=0.4\linewidth, height=0.4\linewidth]{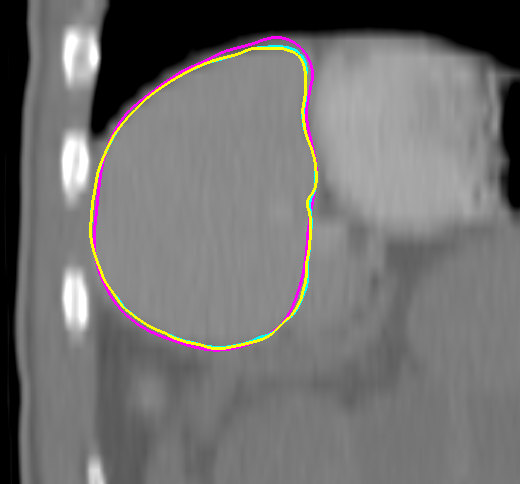}
		\\
		    (c) & (d) \\
	    \end{tabular}

    \caption{Sample spleen segmentation results of proposed methods in the sagittal view. (a) The original image overlaid with the ground truth (purple). (b) The result of V-net (green).  (c) The result with  the proposed CNN only (light blue). (d) The result of the proposed CNN+CRFs (yellow).}

    \label{fig:spleen_good}
\end{figure}
\begin{figure}[h]

    \centering
        \begin{tabular}{c c }
		    \includegraphics[width=0.4\linewidth, height=0.4\linewidth]{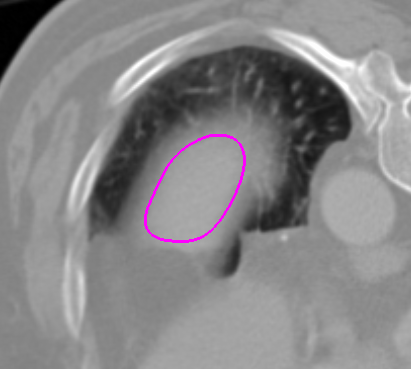} &
		    \includegraphics[width=0.4\linewidth, height=0.4\linewidth]{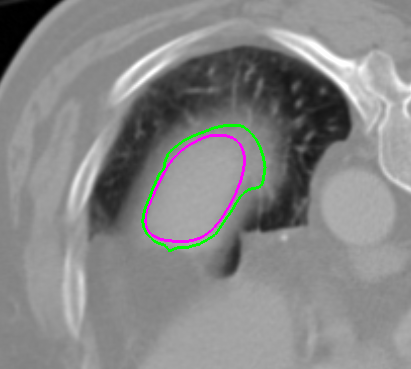} 
		   \\
		 (a) & (b)\\
		    \includegraphics[width=0.4\linewidth, height=0.4\linewidth]{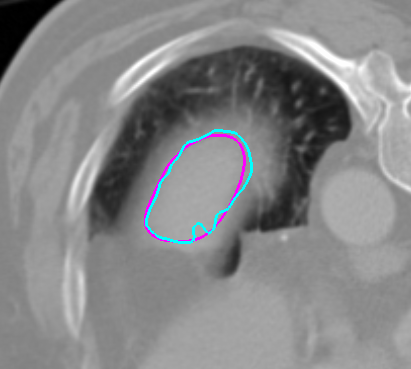} &
		    \includegraphics[width=0.4\linewidth, height=0.4\linewidth]{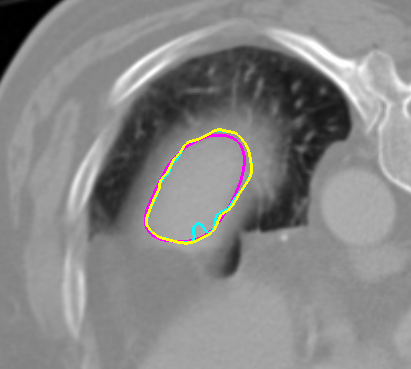}
		\\
		    (c) & (d) \\
	    \end{tabular}

    \caption{A sample spleen slice for which the CRFs layer helps. (a) The original image overlaid with the ground truth (purple). (b) The result of V-net (green).  (c) The result of the proposed CNN (light blue). (d) The result of the proposed CNN+CRFs (yellow).}

    \label{fig:spleen_good2}
\end{figure}

\begin{table}

	\centering
	\caption{Quantitative segmentation results comparison for the MSD Spleen dataset.}

	\begin{tabular}{|c|c|c|c|} 
		\hline
		 & DSC & ASD (mm) & HD (mm) \\ 
		\hline
		V-net \cite{milletari2016v}  & 0.94$\pm$0.03 & 1.2$\pm$1.0 & 16.3$\pm$11.2  \\
		\hline
		proposed CNN+CRFs &  \textbf{0.95$\pm$0.02} &  \textbf{0.86$\pm$0.74} & \textbf{13.6$\pm$12.7} \\
		\hline
		p-value & 0.007 & 0.047 & 0.160 \\
		\hline
	\end{tabular}
	\label{table:overall_spleen}

\end{table}
The quantitative results of the proposed method applied to the MSD Spleen dataset are shown in Table.~\ref{table:overall_spleen}. Although the baseline V-net achieved a promising result, our proposed method is capable of further improving the segmentation accuracy, especially the performance metrics based on the boundary surface distances. Table.~\ref{table:overall_spleen} reveals that, the proposed CNN+CRFs  significantly outperformed V-net with respect to the DSC and ASD in relation to the $p$ values of the $t$-tests. Sample segmentation results are shown in Fig.~\ref{fig:spleen_good}. Illustrative examples showing that the CRFs layer improved the segmentation results can be found in Fig.~\ref{fig:spleen_good2}.

\section{Discussion}
\subsection{Consistent Patches Simplify Task for Networks}
One advantage of the proposed patch generation method is that it computes all consistent patches, i.e. the target surface in every patch is monotonic (each column has exactly one voxel on the surface). By contrast the paradigms of the ground truth for the image patches in the {\em region-based} methods have many variations: the whole patch could be the background or the foreground, or the fraction between background and foreground could vary greatly. The patch consistency in the proposed method may help bring the network to quickly concentrate on the terrain-like surface identification. One may think of it as an example of the attention mechanism \cite{chen2016attention}.  
 
\subsection{Pre-segmentation with a Correct Topology}
The proposed method relies on the fact that the pre-segmentation needs to have the correct topology of the target object, as the computed result always complies with the pre-segmentation. We have experimentally verified that the proposed method is insensitive to the accuracy of the pre-segmentation as long as the topology is correct. With this in regard, model-based methods may work better for pre-segmentation than those that cannot guarantee the topology of the result. For example, a simple U-net/V-net may not the correct (or best) choice for direct use in pre-segmentation as it may not be able to produce the right topology of the target object, although the DSC of its prediction may be significantly higher than that of a simple model-based method. In addition, the pre-segmented surface needs to be sufficiently smooth, as seen in the proposed method where we needed to resample the image volume based on the pre-segmentation. For the spleen dataset, we applied a recursive Gaussian mask smoothing filter and the windowed-sinc filter to smooth the pre-segmented surface. 

\subsection{Inference for Overlapping Patches}
In our current implementation, we compute 6 patches for each image volume with no overlapping between patches. In the {\em region-based} CNN segmentation work, it is commonly known that using overlapped patches and averaging the predictions on the overlapped regions during inference can improve the segmentation results. The same strategy can also be used in our proposed {\em surface-based} segmentation method.

\subsection{Possible Drawbacks of the Proposed Method}
In the {\em region-based} CNN+CRFs framework \cite{zheng2015conditional}, the visual feature is a pixel or voxel intensity of the original image, which is helpful for the CRFs to accurately define the true object boundary to compensate for the coarseness of the semantic segmentation with CNNs. By contrast in our current {\em surface-based} CNN+CRFs framework, the probability map of each column from the CNN part instead of the original image visual feature is fed into the CRFs layer. The probability map generally may include more global information and may lose the exact local boundary information, which may hinder the proposed method’s ability to recover exact surfaces. In GS framework, this problem is remedied by using a carefully designed unary cost term, which includes rich lower level original image information. As part of a future study, we will resolve the problem by integrating information from dedicated local filters or features generated in early stages of CNN. 

Applying the proposed CNN+CRFs framework for the segmentation of medical objects with a very complex structure, (such as nervous structures in the brain MRI, and the airway/vessel tree in the pulmonary CT), may prove challenging. To segment those complex structures, more carefully designed pre-segmentation approaches need to be used. In addition, the sampling directions for image feature columns may also need to be handled carefully such that no two columns interfere with each other. Possible options include the electric field line-based method \cite{yin2009electric} and the generalized gradient vector flow-based method \cite{oguz2014logismos}. 

\subsection{Future Work on Loss Functions}
The MCE loss may not be the best option for our {\em surface-based} segmentation network, as the labels within each column have their ordering. By using the MCE loss, it does not really take the label ordering into account. In the future, we may consider a weighted MCE loss, in which the weight of a label in each column should be proportionally relative to its distance to the ground truth label. Another possible solution is to find out a proper way to optimize the surface position errors, for example by using a mean square error, directly.

\section{Conclusion}
We propose a novel direct surface segmentation method in $3$-D using deep learning. Our approach contrasts with the classification based semantic segmentation, where post processing is needed to obtain the boundary surface of the target object. With the proposed patch generation method, the known topology of the target object can be guaranteed in our segmentation neural network. The CRFs model is seamlessly integrated into the proposed convolutional neural networks for the direct surface segmentation. Thus, the whole CNN+CRFs network can be trained in an end-to-end fashion. The segmentation results tested on the NCI-ISBI 2013 Prostate dataset and the MSD Spleen dataset are promising.

\section*{Acknowledgment}
The research was supported, in part, by US National Science Foundation grant CCF-1733742. This work was done when Z. Zhong and A. Shah were with the Department of Electrical and
Computer Engineering, University of Iowa, Iowa City.


\ifCLASSOPTIONcaptionsoff
  \newpage
\fi


\bibliographystyle{IEEEtran}

\bibliography{ref}
\end{document}